# An optimal method for wake detection in SAR images using Radon transformation combined with wavelet filters

* Ms.M.Krishnaveni   ** Mr. Suresh Kumar Thakur   *** Dr.P.Subashini

*** Research Assistant-NRB, Department of Computer Science, Avinashilingam University for Women, Coimbatore, India

** Deputy Director, Naval Research Board-DRDO, New Delhi, India.

*Lecturer (SG), Department of Computer Science, Avinashilingam University for Women, Coimbatore, India.

*Abstract -* **A new-fangled method for ship wake detection in synthetic aperture radar (SAR) images is explored here. Most of the detection procedure applies the Radon transform as its properties outfit more than any other transformation for the detection purpose. But still it holds problems when the transform is applied to an image with a high level of noise. Here this paper articulates the combination between the radon transformation and the shrinkage methods which increase the mode of wake detection process. The latter shrinkage method with RT maximize the signal to noise ratio hence it leads to most optimal detection of lines in the SAR images. The originality mainly works on the denoising segment of the proposed algorithm. Experimental work outs are carried over both in simulated and real SAR images. The detection process is more adequate with the proposed method and improves better than the conventional methods.**

## I INTRODUCTION

In navy radar applications, the presentation of the radar image traditionally has been the way for the radar operator to interpret the information manually. The large increase in calculation capacity of the image processing in modern radar systems has great effects in detection and extraction of targets[5]. With powerful image processing techniques and algorithms, modern radar systems has the possibility to extract targets and their velocity from the surrounding background. A condition for this automatic detection is that the radar image should be relatively free from undesired signals [2]. Such undesired signals can be rain clutter, sea clutter, measuring noise, landmasses, birds etc. Conventional filtering like Doppler, median and wiener filtering is often used to remove these undesired signals and extract the interesting part of the radar image. Image processing techniques will improve the radar image and investigate an automatic classification and presentation of the objects. The analysis of ship wakes in SAR imagery with specialized algorithms can provide significant information about a wake's associated vessel, including the approximate size and heading of the ship [13]. The velocity of the vessel can be estimated by measuring the disarticulation of the ship relative to the height of the wake. Image processing algorithms such as the Fourier Transform and the Radon Transform allow the user to manipulate SAR images in a way that dramatically increases the chance of detecting ship wakes [2]. The paper is organized as follows: Section 2 deals with the Image localization (SAR images). Section 3 deals with wavelet denoising methods and its metrics. Section 4 comprises the comparison radon transformation and its performance. Section 5 converses the experimental results of the shrinkage methods and radon transformation. This paper also concludes with remarks on achievable prospects in this area.

## II. IMAGE LOCALIZATION

This is the first and lowest level operation to be done on images. The input and the output are both intensity images. The main idea with the preprocessing is to suppress information in the image that is not relevant for its purpose or the following analysis of the image. The pre-processing techniques use the fact that neighboring pixels have essentially the same brightness. There are many different pre-processing methods developed for different purposes. Interesting areas of pre-processing for this work is image filtering for noise suppression. Conservative methods based on wavelet transforms have been emerged for removing Gaussian random noise from images [1]. This local preprocessing speckle reduction technique is necessary prior to the processing of SAR images. Here we identify wavelet Shrinkage or thresholding as denoising method [3]. It is well known that increasing the redundancy of wavelet transforms can significantly improve the denoising performances [7][8].

Thus a thresholding process which passes the coarsest approximation sub-band and attenuates the rest of the sub-bands should decrease the amount of residual noise in the overall signal after the denoising process [4].

### III. IMAGE DENOISING USING WAVELET

The two main confines in image accuracy are categorized as blur and noise. Blur is intrinsic to image acquisition systems, as digital images have a finite number of samples and must respect the sampling conditions. The second main image perturbation is noise. Image denoising is used to remove the additive noise while retaining as much as possible the important signal features[1]. Currently a reasonable amount of research is done on wavelet thresholding and threshold selection for signal de-noising, because wavelet provides an appropriate basis for separating noisy signal from the image signal[3]. Two shrinkage methods are used over here to calculate new pixel values in a local neighborhood. Shrinkage is a well known and appealing denoising technique[9][10]. The use of shrinkage is known to be optimal for Gaussian white noise, provided that the sparsity on the signal's representation is enforced using a unitary transform[6]. Here a new approach to image denoising, based on the image-domain minimization of an estimate of the mean squared error-Stein's unbiased risk estimate (SURE) is proposed and equation (1) specifies the same. Surelet the method directly parameterizes the denoising process as a sum of elementary nonlinear processes with unknown weights. Unlike most existing denoising algorithms, using the SURE makes it needless to hypothesize a statistical model for the noiseless image. A key of it is, although the (nonlinear) processing is performed in a transformed domain-typically, an undecimated discrete wavelet transform, but we also address nonorthonormal transforms-this minimization is performed in the image domain [6].

$$sure(t;x) = d - 2.\#\{i : |x_i| \leq t\} + \sum_{i=1}^{d}(|x_i|\Lambda t)^2 \text{ --(1)}$$

where *d* is the number of elements in the noisy data vector and *xi* are the wavelet coefficients. This procedure is smoothness-adaptive, meaning that it is suitable for denoising a wide range of functions from those that have many jumps to those that are essentially smooth.

It have high characteristics as it out performs Neigh shrink method. Comparison is done over these two methods to prove the elevated need of Surelet shrinkage for the denoising the SAR images. The experimental results are projected in graph format which shows that the Surelet shrinkage minimizes the objective function the fastest, while being as cheap as neighshrink method[15]. Measuring the amount of noise equation (2) is done by its standard deviation , $\sigma(n)$, one can define the signal to noise ratio (SNR) as

$$SNR = \frac{\sigma(\mu)}{\sigma(n)}, \quad \text{-- (2)}$$

Where $\sigma(\mu)$ in equation (3) denotes the empirical standard deviation of $\mu(i)$,

$$\sigma(\mu) = \left(\frac{1}{|I|}\sum_{i}(u(i) - \overline{\mu})^2\right)^{1/2} \text{--(3)}$$

And $\overline{\mu} = \frac{1}{|I|}\sum_{i \in I}\mu(i)$ is the average grey level value.

The standard deviation of the noise can also be obtained as an empirical measurement or formally computed when the noise model and parameters are known. This parameter measures the degree of filtering applied to the image [5]. It also demonstrates the PSNR rises faster using the proposed method than the former. Hence the resulted denoised image is conceded to the next segment for the transformation to be applied and it is also proved to improve detection process.

### IV. RADON TRANSFORMATION

Detection of ships and estimating their velocities are major work done in SAR images. Here the proposed method takes advantage of two thresholding techniques and inserts some innovation by using the Radon Transform to detect the ship wake and estimate the range velocity component[12]. The proposed technique was applied to synthetic raw data, which contains a moving vessel and its respective wake. The Radon Transform calculates the angle that a straight line perpendicular to the track makes with the x-axis in the center of the image. Knowing this, simply add 90º to the value obtained to find the angle of the wake arm. If an image is consider as *I*, with dimensions MxM. The Radon transform $\hat{I}$ is given in equation (4)

$$\hat{I}(\chi_\theta, \theta) = \sum_{y\theta=-M/2}^{M/2} I(x_\theta\cos\theta - y_\theta\sin\theta, x_\theta\sin\theta + y_\theta\cos\theta) \text{-(4)}$$

where $((x_\theta, y_\theta) \in Z$ and $\theta \in [0; \pi]$

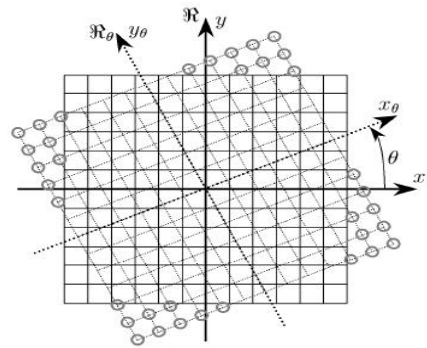

Several definitions of the radon transform exists, and expresses lines in the form of rho=x*cos(theta)+y*sin(theta), where theta is the angle and rho the smallest distance to the origin of the coordinate system[12]. The Radon transform for a set of parameters (rho,theta) is the line integral through the image g(x,y), where the line is positioned corresponding to the value of (rho,theta). The delta() is the Dirac delta function which is infinite for argument 0 and zero for all other arguments[14].

This function is implied with the original image, and denoised image of two methods[11]. The detection of line segment in the SAR images is more appropriate with surelet denoising and radon transformation then with the former and the conventional method. Experiments are carried over with the proposed method to verify and validate the results. With the angle of both arms of the wake calculated, the equation of the line that passes by each of them can be estimated.

## V. RESULTS AND DISCUSSION

To verify the validity of the proposed method the results are compared based on PSNR ratio and time parameters for the Shrinkage methods and it is given in figure1. With the extension of the next segment work, detection of angle is also compared based on the radial coordinates (rho). Noise (sigma) is been the main phenomena for the comparison job. Surelet which is the latest method based on the SURE. The DWT was used with Daubechies, least asymmetric compactly-supported wavelet with eight vanishing moments with four scales. The 120 x 120 pixel region SAR images are used for applying radon transformation. They were contaminated with Gaussian random noise of 10 20 30 50 75 100.

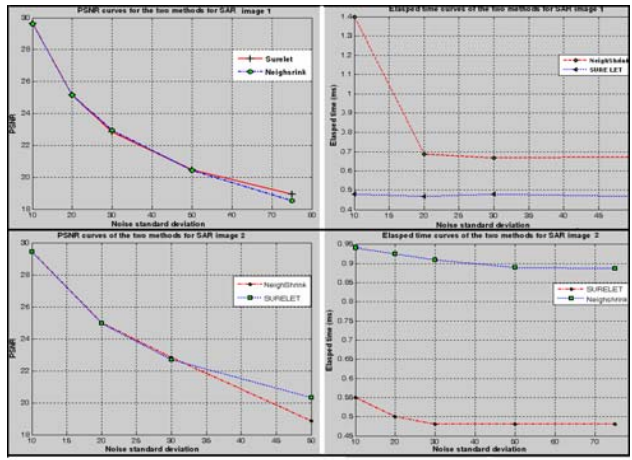

**Figure 1: Comparison of PSNR values and time for two Methods (NeighShrink and surelet) for Two SAR image**

For the wake detection the angle is got by applying the radon transformation which in results the same angle value with variations in rho values which is shown in figure2.

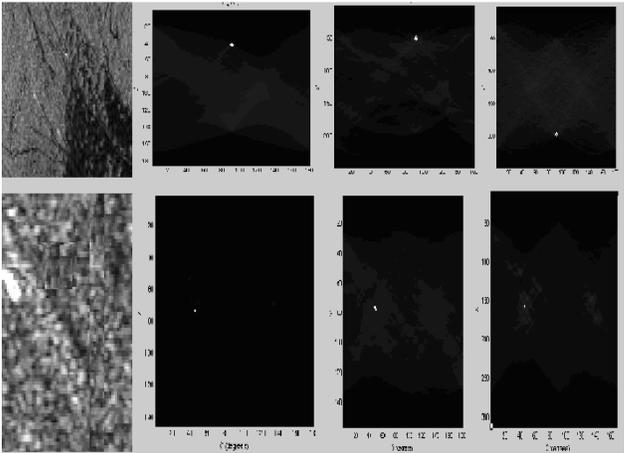

**Figure 2: (a) Original image (b) Angle using RT
(c) Angle using first denoising method and RT
(d) Angle using second denoising method and RT**

Table 1 explicates about the radial co ordinates values and the angle values for two SAR images with corresponding change of noise values for each method respectively.

| SAR images | Noise values | Original Image with RT | | Denoised image (first method with RT) | | Denoised image (second method with RT) | |
|---|---|---|---|---|---|---|---|
| | | radial | angle | radial | angle | radial | angle |
| Image 1 | 10 - 100 | 48 | 85 | 50 | 85 | 195 | 85 |
| Image 2 | 10 - 100 | 85 | 45 | 85 | 45 | 155 | 45 |

**Table 1: Comparison of three methods with change of noise values**

## VI. CONCLUSION

In this proposed method, the originality of the technique consent to the wake detection and the estimation of the velocity of vessels more effectively. Here the projected method proves that surelet compared with Neighshrink can determine optimal results by using finest threshold instead of using the suboptimal universal threshold in all bands. It exhibits an excellent performance for wake detection and the experimental result signifies that it produces both higher PSNRs and enhanced visual eminence than the former and conventional methods. The Radon Transform is used to detect the ship wake and estimate the range velocity component. The key advantage is that it holds low computational requirements. Further enhancement of the work can be concentrated on the neighbouring window size for every wavelet sub-band .This aid in difficulties when the ship wake is not visible in the image properly. This paper is therefore concluded that better detection with lower probability of false alarm rate.